\documentclass[10pt,twocolumn,letterpaper]{article}

\usepackage{cvpr}              

\definecolor{cvprblue}{rgb}{0.21,0.49,0.74}
\usepackage[pagebackref,breaklinks,colorlinks,allcolors=cvprblue]{hyperref}

\usepackage{algorithm}
\usepackage{algorithmicx}
\usepackage{algpseudocode}
\usepackage{amsmath}
\usepackage{multirow}
\usepackage{makecell}
\usepackage{bm}

\def\paperID{15336} 
\def\confName{CVPR}
\def\confYear{2025}

\title{MoManipVLA: Transferring Vision-language-action Models for General Mobile Manipulation}

\author{Zhenyu Wu\textsuperscript{1} \quad 
Yuheng Zhou\textsuperscript{2} \quad
Xiuwei Xu\textsuperscript{3} \quad
Ziwei Wang\textsuperscript{2} \quad
Haibin Yan\textsuperscript{1$*$} \quad \\
	\textsuperscript{1}Beijing University of Posts and Telecommunications \qquad \textsuperscript{2}Nanyang Technological University \\  \qquad \textsuperscript{3}Tsinghua University\\ 
	{\tt\small\{wuzhenyu, eyanhaibin\}@bupt.edu.cn  \quad zhou0484@e.ntu.edu.sg} \\
	{\tt\small xxw21@mails.tsinghua.edu.cn \quad ziwei.wang@ntu.edu.sg}}

\begin{document}
\twocolumn[{%
\renewcommand\twocolumn[1][]{#1}%
\maketitle

\begin{center}
    \centering
    \includegraphics[width=1.0\linewidth]{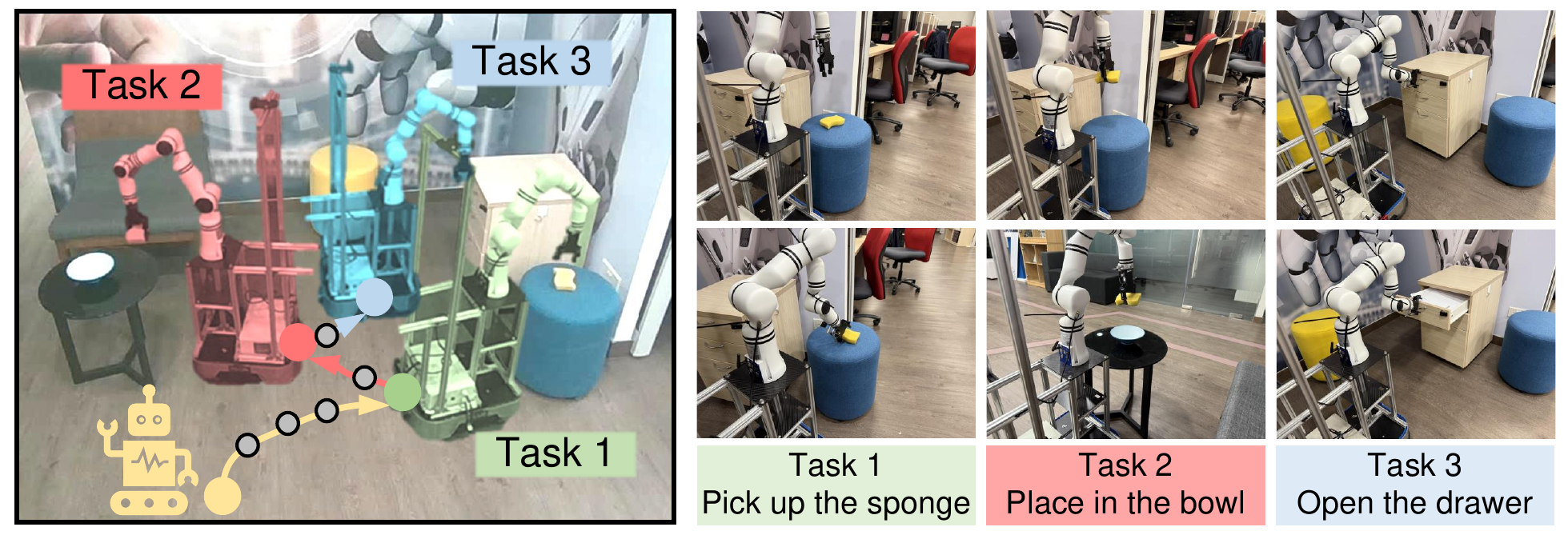}
    \captionof{figure}{Transferring pre-trained VLA models to mobile manipulation significantly enhances the generalization ability of the policy across tasks and environments. Our MoManipVLA can complete diverse household tasks such as object picking, object delivery and drawer opening in large working area. Guided by the waypoints from VLA models, the motions of the mobile base and the robot arm are jointly generated with physical feasibility constraints.}
   \label{fig:compare}
\end{center}
}]



\begin{abstract}
Mobile manipulation is the fundamental challenge for robotics to assist humans with diverse tasks and environments in everyday life. 
However, conventional mobile manipulation approaches often struggle to generalize across different tasks and environments because of the lack of large-scale training.
In contrast, recent advances in vision-language-action (VLA) models have shown impressive generalization capabilities, but these foundation models are developed for fixed-base manipulation tasks.
Therefore, we propose an efficient policy adaptation framework named MoManipVLA to transfer pre-trained VLA models of fix-base manipulation to mobile manipulation, so that high generalization ability across tasks and environments can be achieved in mobile manipulation policy.
Specifically, we utilize pre-trained VLA models to generate waypoints of the end-effector with high generalization ability. We design motion planning objectives for the mobile base and the robot arm, which aim at maximizing the physical feasibility of the trajectory. Finally, we present an efficient bi-level objective optimization framework for trajectory generation, where the upper-level optimization predicts waypoints for base movement to enhance the manipulator policy space, and the lower-level optimization selects the optimal end-effector trajectory to complete the manipulation task.
In this way, MoManipVLA can adjust the position of the robot base in a zero-shot manner, thus making the waypoints predicted from the fixed-base VLA models feasible.
Extensive experimental results on OVMM and the real world demonstrate that MoManipVLA achieves a 4.2\% higher success rate than the state-of-the-art mobile manipulation, and only requires 50 training cost for real world deployment due to the strong generalization ability in the pre-trained VLA models.
Our project homepage can be found at \href{https://gary3410.github.io/momanipVLA/}{here}.

\end{abstract}    
\section{Introduction}
\label{sec:intro}
Mobile manipulation empowers robotics with the ability to perform complex manipulation tasks across large spaces, which requires the whole-body control for the mobile base and the arm \cite{xiong2024adaptive, yang2023harmonic}.
With the increasing popularity of intelligent robotic systems, several fields such as household service \cite{xiao2024robi}, manufacturing \cite{peng2024revolutionizing}, and logistics \cite{vstibinger2021mobile} are in urgent requirement of mobile manipulation capabilities since robotics are needed to perform cross-space manipulations autonomously. 
However, the requirement to perform diverse tasks in unstructured environments (e.g., assist humans in their daily lives) presents significant challenges.

Conventional mobile manipulation frameworks separately train navigation and fixed-base manipulation modules, leading to compounding errors \cite{yin2024sg, choi2024find, fang2023anygrasp, ze2023gnfactor, gu2022multi}. Recent end-to-end methods jointly optimize navigation and manipulation actions, but high collection costs for demonstrations limit dataset scale and generalization \cite{huang2023skill, yokoyama2023asc}. Meanwhile, although VLA models show strong generalization in diverse manipulation tasks \cite{li2024manipllm, kim2024openvla}, their focus on fixed-base tasks prevents them from generating cooperative actions between the mobile base and robot arm for mobile manipulation.

In this paper, we propose an efficient policy transfer framework named MoManipVLA to generalize the fixed-base VLA models to mobile manipulation tasks. 
Unlike existing mobile manipulation methods which suffer from low generalization ability across tasks and environments, MoManipVLA efficiently transfers generalizable fix-base manipulation policy from pre-trained VLA models to mobile manipulation.
More specifically, we employ pre-trained VLA models to predict end-effector waypoints with high generalization ability to guide the generation of mobile manipulation trajectories.
We then design motion planning objectives of scene constraints for the mobile base and the robot arm including end-effector reachability, trajectory smoothness and collision avoidance, which aim to maximize the physical feasibility of the trajectories. 
To efficiently plan the whole-body motion, we propose a bi-level trajectory optimization framework of the objectives, where the upper-level optimization predicts the waypoints for base movement to strengthen the manipulator policy space, and the lower-level optimization selects the optimal end-effector trajectory for task completion.
Figure \ref{fig:compare} illustrates the comparison between the conventional and the proposed approaches, {MoManipVLA efficiently transfers the pre-trained VLA models to diverse mobile manipulation tasks, where the mobile base and the robot arm coordinately perform actions with physically feasible trajectories.
Extensive experimental results on OVMM \cite{yenamandra2023homerobot} and real world demonstrate that our method achieves a 4.2\% higher success rate than state-of-the-art mobile manipulation techniques. 
Due to the strong generalization of our pre-trained VLA models, MoManipVLA only requires 50 expert episodes for real-world deployment.
Our contributions include:

\begin{itemize}
\item We propose a policy adaptation framework that transfers pre-trained VLA models for mobile manipulation, enabling high generalization across tasks and environments.
\item We introduce motion planning objectives and a bi-level trajectory optimization framework to enhance the physical feasibility and efficiency of trajectories.
\item Extensive experiments in OVMM and the real world validate the method's generalization and efficiency.
\end{itemize}
\section{Related Work}
\label{sec:formatting}

\subsection{Mobile Manipulation Framework}
Mobile manipulation requires agents to have the ability to interact with objects in large spaces based on human instructions. 
Existing mobile manipulation frameworks can be categorized into two types: end-to-end and modular. 
End-to-end approaches \cite{brohan2022rt, gupta2018robot, fu2024mobile} employ imitation learning to directly predict mobile manipulation actions based on visual observations.
Qiu \emph{et al.} \cite{qiu2024learning} propose to learn a unified scene information representation for navigation and manipulation, which leverages both geometric and semantic information to improve the success rate of the manipulation.
Yan \emph{et al.} \cite{yan2024m2diffuser} utilized diffusion policy to generate mobile manipulation whole-body control trajectories, which control task-specific diffusion policy generation through diverse energy terms.
However, imitation learning with expert trajectories leads to expensive training costs \cite{mendonca2024continuously}.
Therefore, modular mobile manipulation frameworks Home-Robot \cite{yenamandra2023homerobot} and OK-Robot \cite{liu2024ok}, which contain foundation model planners and reinforcement learning-based controllers are proposed to efficiently handle long-horizon mobile manipulation tasks.
SPIN \cite{uppal2024spin} proposes to utilize a reactive mobile manipulation framework to achieve proactive scene perception which follows the full-body and hand-eye coordination capabilities of humans.
However, existing mobile manipulation methods struggle to generalize to diverse real-world tasks and environments due to insufficient large-scale pre-training.

\subsection{Vision-Language-Action Model}
The current state-of-the-art VLA models \cite{kim2024openvla, li2024manipllm, huang2023embodied, li2023vision} directly output end-effector 7-DoF actions based on RGB vision observations without relying on predicted object categories and poses. 
Early VLA frameworks following the vision-language model(VLM) architecture represented actions with autoregressive discretization.
ManipLLM \cite{li2024manipllm} constructed the Chain-of-Thought(CoT) to stimulate manipulation reasoning capabilities in the foundation model and further introduced an active impedance adaptive strategy to plan the next waypoints.
OpenVLA \cite{kim2024openvla} explored the impact of visual encoders on the performance of the VLA models, which combines different foundation model components to achieve satisfactory performance.
To further enhance the VLA model to mine the association between visual input and action trajectories, TinyVLA \cite{wen2024tinyvla} proposes to utilize the foundation model feature prior to guide the action decoder diffusion process, which significantly enhances the generalization of the VLA model on changes in viewpoints, objects, etc.
Recent works also explore the ability of VLA models on embodiments with high degrees of freedom.
RDT-1B \cite{liu2024rdt} generalized diffusion policy to a bimanual manipulation by introducing a physically interpretable unified action space to handling more complex tasks.
$\pi$0 \cite{black2024pi_0} proposed a novel flow-matching architecture based on VLM to learn internet-scale knowledge, which empowered cross-embodied manipulation planning for VLA models. 
Although existing VLA models demonstrate impressive generalization across tasks and environments, they fail to generate the cooperative actions of the mobile base and the robot arm.

\subsection{Trajectory Optimization}
In robotic manipulation tasks, trajectory optimization plays a pivotal role in achieving efficient and precise robot actions.
Early research has focused on the multi-objective and constraint characteristics of trajectory optimization \cite{kelly2017introduction, hewing2020learning, chang2016compositional}, aiming to address the demands for accuracy and low latency within dynamic environments.
Handcrafted optimization objectives confine conventional methods to specific tasks, hindering their deployment in dynamic and complex environments. 
Consequently, data-driven approaches are now used to tailor trajectory optimization to the dynamic challenges of real-world scenes.
Contact-GraspNet \cite{sundermeyer2021contact} and O2O-Afford \cite{mo2022o2o} guide trajectory generation in densely cluttered scenes by predicting grasp poses and object affordances, respectively. 
Michael \emph{et al.} \cite{danielczuk2021object} developed a collision model that predicts 6DOF object constraints, effectively tackling challenges associated with occlusions in point cloud data.
To leverage the knowledge embedded in the demo episodes, some works employ imitation learning to mine trajectory policies directly from expert demonstrations.
Chi \emph{et al.} \cite{chi2023diffusion} and Huange \emph{et al.} \cite{huang2023diffusion} train neural networks to generate collision-free trajectories through imitation learning from human or expert demonstrations, eliminating the need for manually defined constraints or explicit environmental modeling. 
Benefit from the development of foundation models, recent methods such as VoxPoser \cite{huang2023voxposer} and ReKep \cite{huang2024rekep} have demonstrated that leveraging foundational models to infer scene affordances and constraints can significantly enhance the generalization of trajectory optimization. 
Inspired by Rekep’s success in manipulation tasks, we developed a motion planning objective that efficiently adapts pre-trained fixed-base VLA models for mobile manipulation.

\begin{figure*}[t]
  \centering
    \includegraphics[width=1.01\linewidth]{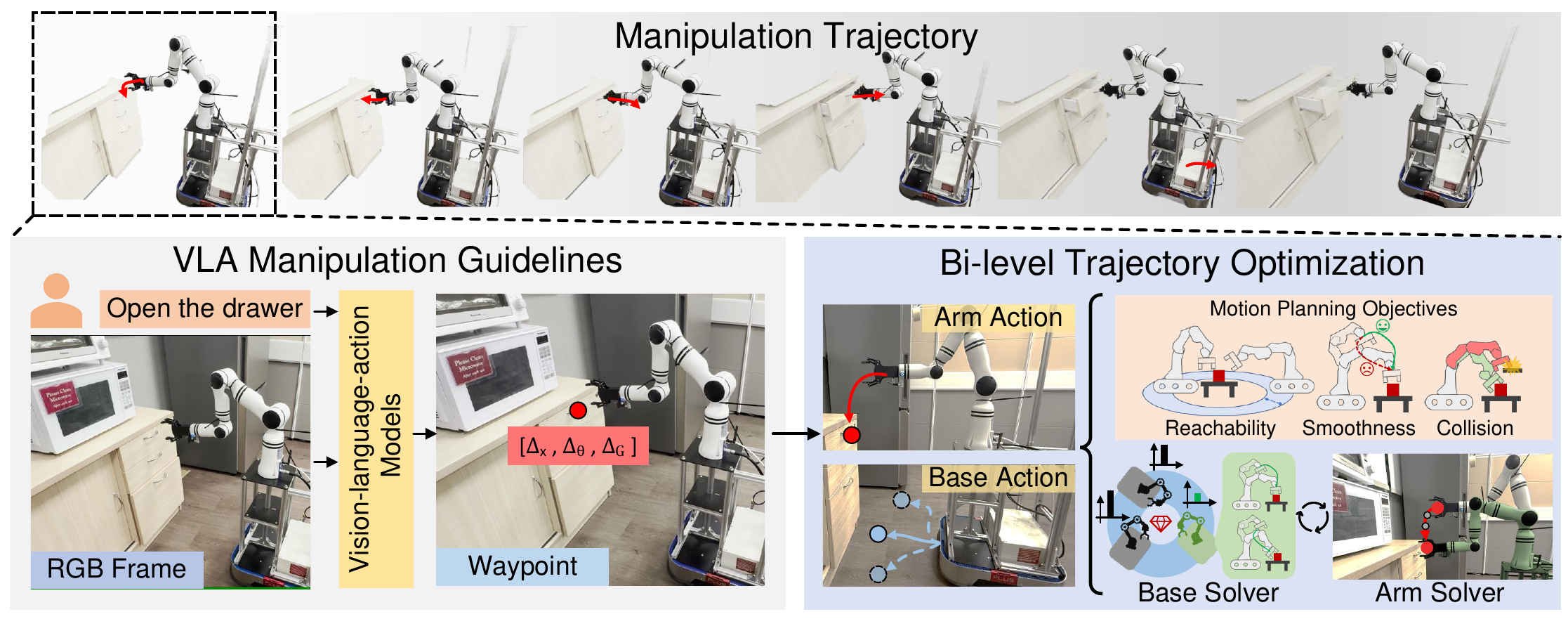}

   \caption{The pipeline of MoManipVLA. The pre-trained VLA models predict highly generalized end-effector waypoints to guide the mobile manipulation task, through which the trajectory of the mobile base and the robot arm can be generated with objectives of physical feasibility. The objectives consider the reachability, smoothness and collision, and the trajectory is acquired via bi-level optimization.}
   \label{fig:pipeline}
   \vspace{-0.3cm}
\end{figure*}

\section{Approach}
In this section, we first introduce the overall framework of policy transfer from the fixed-base VLA models to mobile manipulation tasks. Then we present the motion planning objectives for evaluation of the base and the arm trajectory to maximize the physical feasibility. Finally, we propose an efficient bi-level objective optimization framework for trajectory generation.

\subsection{Problem Statement}
The mobile manipulation task requires the robot to interact with objects in different locations, cooperatively controlling the mobile base and the robot arm to accomplish the interaction actions in physically feasible trajectories. 
The observation at the $t_{th}$ step consists of the visual input and the proprioception state. The visual input contains RGB images, depth images, and the camera poses, and the proprioception state includes locations $p_b^t$ and rotations $r_b^t$ for the base in the world frame, and locations $p_e^t$, rotations $r_e^t$, and gripper openness $g_e^t$ of the end-effector in the base frame.
We apply the transformation function $\Gamma$ to obtain the location $\hat{p}_{e}^{t}$ and rotation $\hat{r}_{e}^t$ of the end-effector in the world frame.
Based on observation and language instructions, the agent is required to generate the optimal action to achieve the next-best proprioception state of the mobile base and the robot arm for task completion. 

Existing mobile manipulation frameworks usually learn the mobile manipulation policy by imitating expert demonstrations. The scale of expert demonstrations is limited because of the high cost of data collection in mobile manipulators with high DoFs, which leads to low generalization ability across tasks and environments. Our goal is to efficiently transfer pre-trained VLA model policies to mobile manipulation, which jointly generate trajectories of both the mobile base and the robot arm to achieve high generalization ability.



\subsection{Policy Transfer Framework}
VLA models pre-trained on web-scale data \cite{o2023open} demonstrate impressive generalization capabilities across tasks and environments. 
Nevertheless, existing efforts are limited to fixed-base manipulation and cannot deal with mobile manipulation that requires whole-body policy generation for both the base and the arm.
To this end, we propose to transfer the fixed-base manipulation policy predicted by pre-trained VLA models by whole-body motion planning with physical feasibility. Figure 2 demonstrates the overall pipeline of MoManipVLA.
We first employ pre-trained VLA models to generate the optimal end-effector waypoints based on observation and human instruction. 
To enable the end-effector to achieve the goal waypoints, we jointly generate the trajectories for the base and the arm with the highest physical feasibility. We design motion planning objectives to maximize the physical feasibility including safety, smoothness and reachability.
To efficiently solve the trajectory, we propose a bi-level objective optimization framework, where the upper-level optimization predicts the base movement waypoints to enhance the subsequent manipulation policy space, and the lower-level optimization selects the optimal end-effector trajectory to achieve the waypoints generated by VLA models. Therefore, the motion planning framework enables pre-trained fixed-base policies to be efficiently adapted to mobile manipulation with negligible training cost.

\subsection{Mobile Motion Planning Objectives}
Motion planning aims to generate feasible trajectories for the mobile base and the robot arm between given the consecutive waypoints of end-effectors, which should achieve pose reachability, trajectory smoothness and safety during the entire mobile manipulation process.
To enhance the generalization ability across tasks and environments, the $i_{th}$ waypoint of the end-effector $\mathbf{Q}_{i}$ in the base frame during the entire mobile manipulation process can be generated via a pre-trained fixed-base VLA model, where RGB images of the current scene and the proprioception state of the robot arm are considered folllowing the VLA pipeline.
For trajectory planning, we first transform the waypoint $\mathbf{Q}_{i}$ to that the world frame denoted as $\hat{\mathbf{Q}}_{i}$ in the following:
\begin{equation}
    \hat{\mathbf{Q}}_{i} = \Gamma(\mathbf{P}_{i}^0, \mathbf{Q}_{i})
\end{equation}
where $\mathbf{P}_{i}^0$ demonstrates the proprioception state of the base when predicting the waypoint $\mathbf{Q}_{i}$ with the VLA model. Therefore, the optimization objective can be represented as follows to generate the trajectory given the consecutive waypoints $\hat{\mathbf{Q}}_{i}$ and $\hat{\mathbf{Q}}_{i+1}$:
\vspace{-2mm}
\begin{equation}
\begin{aligned}
&\min \sum_{t=0}^{T_i}\mathcal{O}(\mathbf{x}_{i,b}^t, \mathbf{x}_{i,e}^t) \\ 
&\text{s.t.}quad \Gamma(\mathbf{x}_{i,b}^0, \mathbf{x}_{i,e}^0) = \hat{\mathbf{Q}}_{i}, \ \quad \Gamma(\mathbf{x}_{i,b}^{T_i}, \mathbf{x}_{i,e}^{T_i}) = \hat{\mathbf{Q}}_{i+1} \\
\end{aligned}
\end{equation}
where $T_i$ denotes the number of steps in the planned trajectory between above consecutive waypoints, and $\mathcal{O}$ is the objective function evaluating the physical feasibility of trajectories. For trajectory planning, $\mathbf{x}_{i,b}^t$ means the planned base pose in the world frame at the $t_{th}$ step including location $p_{i,b}^t$ and rotation $r_{i,b}^t$, and $\mathbf{x}_{i,e}^t$ demonstrates the planned end-effector pose in the base frame at the $t_{th}$ step including location $p_{i,e}^t$, rotation $r_{i,e}^t$ and gripper openness $g_{i,e}^t$. 
Our goal is to minimize the cost function to enhance the physical feasibility of the entire trajectory with waypoints from VLA models, and the constraints indicate that the starting pose and the ending pose should be consistent with the waypoints for trajectory smoothness.
Given the poses of the base and the arm in the trajectory, the joint angles of the arm are iteratively solved by Pinocchio IK solver, and the base is driven with translation and rotations. The objective functions are introduced in the following.

\textbf{Reachability cost:}
Since the mobile manipulation requires the robot to interact with objects in large area, improper pose of the mobile base may lead to unreachable pose of the arm. Therefore, we evaluate the reachability of each candidate trajectory. Assuming the solution of joint angles can be acquired via $N_{IK}$ iterations, the reachability cost $\mathcal{F}_{r}$ can be represented as follows:
\begin{equation}
 \mathcal{F}_{r} = 
    \left\{
    \begin{aligned} 
        & N_{IK}/N_{max}  &&\text{if}~~ N_{IK} \leqslant N_{max}, \\
        & C_0 \quad &&\text{if}~~ N_{IK} > N_{max},
    \end{aligned}
    \right.
\end{equation}
where $N_{max}$ is a hyperparameter representing the maximum iteration time for the IK solver, and $C_0$ is a large constant. Acquiring the joint angle solution within the maximum iterations indicates reachable trajectories. Slower IK solving with more iterations means that the joint angle is closer to the range limit, which is more vunerable to noise in execution the mobile manipulation actions. This shows lower reachability for the robot arm. When the iteration number is larger than the budget, the pose in the candidate trajectory is unreachable without feasible joint angle solution. Therefore, the cost will be assigned to a extremely high value.

\textbf{Smoothness cost:}
The smoothness constrains keeps continuous and smooth changes for joint angles of the robot arm and translation and rotation of the base, where sudden changes are avoided to keep the safety of the motor and the embodiment.
We define trajectory smoothness as the difference of joint angles $\bm{\theta}^{t}$ of the arm and proprioception of the base between consecutive poses in the candidate trajectory:
\begin{equation}
    \mathcal{F}_{s} = \sum_{t=0}^{T_i}\left \|\bm{\theta}^{t+1} - \bm{\theta}^{t}  \right \|_2 + \left \|\bm{x}_{b}^{t+1} - \bm{x}_{b}^{t}  \right \|_2
\end{equation}where subscript $i$ for the base pose in the trajectory is omitted for simplicity. We leverage the joint angle solved via IK for smoothness constraint instead of the proprioception of the arm, because small changes in the arm pose cannot guarantee slight difference in joint angles.

\textbf{Collision cost:}
The robot is required to avoid any collision among the robot arm, the mobile base and objects in the environment to keep safety during the mobile manipulation process. We leverage nvblox \cite{millane2024nvblox} to compute the ESDF for object surface in the environment according to the RGB-D images, and randomly sample $N_q$ query points on the robot surface to evaluate the collision cost:
\begin{equation}
    \mathcal{F}_{c} = \sum_{t=0}^T\sum_{j=1}^{N_q} \max(0, ~\epsilon_0-\mathcal{D}(q_j^t, \Omega))  
\end{equation}where $q_j^t$ means the $j_{th}$ query point at the $t_{th}$ pose in the trajectory, and $\epsilon_0$ is a hyperparameter that controls the safety margin for collision avoidance. $\mathcal{D}(q_j^t, \Omega)$ means the distance between $q_j^t$ and the surface $\Omega$, which is predicted by the ESDF model. 
We expect the distance between the robot and the object surface can be maximized if it is within the safety margin. Otherwise, the trajectory will be regarded as collision-free with no contribution to the cost.

We acquire the overall objectives by combining the reachability, smoothness and collision costs with the hyperparameter $\{\lambda_i\}_{i}$:
\begin{equation}
    \begin{aligned}
        \mathcal{O} &= \lambda_{1}\mathcal{F}_{r}+\lambda_{2}\mathcal{F}_{s}+\lambda_{3}\mathcal{F}_{c} \\
    \end{aligned}
    \label{eq:objective}
\end{equation}
By generating physically feasible trajectories, the agent can fully leverage the generalization ability across tasks and environments in VLA models for the challenging mobile manipulation task. 

\begin{algorithm}[t]
\caption{Bi-Level Trajectory Optimization}
\begin{algorithmic}[1]
\Require Solution tolerance: \( \mu_s \)
\Require Upper and lower iterations: $N_{max}^{up}$, $N_{min}^{low}$
\State Get $i_{th}$ waypoint $\hat{\mathbf{Q}}_{i} \leftarrow \Gamma(x_b^0, \mathbf{Q}_{i})$
\State Get initial solution $\{x_e^{0}, x_b^{0}\}$ by linear interpolation
\State  Initialize current stage step $t \leftarrow 0$
\Repeat
\State Update robotic state $\{p_b^t, r_b^t\}, \{p_e^t, r_e^t\}
\leftarrow x_b^t, x_e^t$
\For{$k = 1$ to $N_{max}^{up}$} \Comment{Update Base}
\State Transform end-effector position $\hat{p}_{e}^{t} \leftarrow \Gamma(p_b^{t}, p_e^t)$
\State Sampling end-effector trajectories $L_s$ via $\hat{p}_{e}^t$
\State Update robotic base $x_b^{t+1}$ via Ep.\ref{eq:objective} and $L_s$
\State Update current stage step $t \leftarrow t+1$
\EndFor
\For{$k = 1$ to $N_{max}^{low}$}  \Comment{Update End Effector}
\State Transform end-effector position $\hat{p}_{e}^{t} \leftarrow \Gamma(p_b^{t}, p_e^t)$
\State Update robotic base position $p_e^{t+1}$ via Ep.\ref{eq:objective}
\State Update robotic end-effector $x_{b}^{t+1} \leftarrow x_{b}^t$
\State Update current stage step $t \leftarrow t+1$
\EndFor
\Until{$ \|\Gamma(p_b^{t}, p_e^t) - \hat{\mathbf{Q}}_{i}\|_2 < \mu_{s} $}
\end{algorithmic}
\end{algorithm}

\subsection{Bi-Level Trajectory Optimization}
Directly searching the optimal solution to the objective (\ref{eq:objective}) is highly complex, because the pose search space of the mobile base and the robot arm is extremely large with high DoFs. To address this problem, we propose a bi-level trajectory optimization framework to improve the efficiency of trajectory generation, where the upper level optimizes the waypoints for the base to enhance the manipulator policy space, while the lower level further optimizes the waypoints for the end-effector to follow the guidance of the pre-trained VLA models for completing manipulation tasks.
Although the bi-level optimization framework is greedy, but mobile manipulation requires searching a 10-DoF space (7 for the arm and 3 for the base), resulting in a large search space that cannot be proven convex, where direct searching may easily lead to local optima.
Notably, decomposed greedy search is commonly used in other robotics fields, such as humanoid control [44], which splits the 39-DoF humanoid actions into separate 27-DoF upper and 12-DoF lower body actions.
We employ Dual Annealing search algorithms for the objective optimization, where gradient-based local optimizer SLSQP is used to refine the solution.
The details of bi-level trajectory optimization are demonstrated in Algorithm 1. 

\textbf{Initialization: }According to the constraints in the objective, the first and the last poses of the trajectory are set to the current waypoint and the next one predicted by VLA models. The intermediate poses in the trajectory is initialized as the interpolation states with equal intervals between the consecutive waypoints.

\textbf{Upper-level optimization:} In the upper-level, we search the optimal base poses to form the search space of arm poses. 
In the iterative updating stage of the Dual Annealing search algorithms, we only update the base pose to generate new candidates of whole-body pose in the trajectory. For the $j_{th}$ pose in the trajectory, we randomly sample different arm poses given the base pose, and evaluate the quality of the search space with the expected objective and top-tier objective:
\begin{equation}
    \begin{aligned}
        J_{up} = \sum_{x_e\in \mathcal{M}}\mathcal{O}(x_b^j, x_e)+\alpha \sum_{x_e\in \mathcal{N}}\mathcal{O}(x_b^j, x_e) \\
    \end{aligned}
    \label{eq:upper-level}
\end{equation}where $x_e$ means the candidate arm pose. $\mathcal{M}$ and $\mathcal{N}$ respectively mean the set of sampled candidates and those with the top-k lowest objectives. The search space with low expected objectives across candidates indicates high quality, and lower top-tier objective means the higher upper bound for the candidate performance.

\textbf{Lower-level optimization: }After the base poses in the trajectory achieve optimal, the low-level optimization searches for the best arm poses in trajectory. Given the search space, the quality of each candidate can be estimated by the objective (\ref{eq:objective}). We leverage the optimal candidate in the given search space as the whole-body pose in the trajectory. With the efficient policy transfer, our agent achieve high generalization ability across tasks and environments with the guidance from pre-trained VLA models. 

\section{Experiment}
In this section, we conduct comprehensive experiments in OVMM and the real world to demonstrate the effectiveness of the proposed method. We first describe the implementation details and compare them with state-of-the-art methods. Then we validate the effectiveness of each cost term in the motion planning objective through ablation experiments. Finally, we test the method performance in the real world to verify generalization.

\begin{table*}[t]
\caption{Comparison results on the OVMM benchmark, where partial success rates indicate the execution of each stage. We follow the OVMM setting employing the baseline Nav and Gaze model to find the target object, replacing the Pick and Place policy to validate the effectiveness of the proposed method on mobile manipulation.}
\centering
\setlength{\tabcolsep}{15pt}
\begin{tabular}{@{}lccccccc@{}}
\toprule
\multirow{2}{*}{\textbf{Method}}
& & \multicolumn{3}{c}{\textbf{Partial Success Rates}} &\multirow{2}{*}{\textbf{\makecell{Overall
\\ SR}}} &\multirow{2}{*}{\textbf{\makecell{Partial
\\ SR}}} &\multirow{2}{*}{\textbf{Step}} \\
\cmidrule{3-5} 
  & & FindObj & Pick & FindRex  \\
 \midrule
UniTeam &  &49.2\% &42.8\%  & 19.6\% &  9.2\% & 30.8\% & 1006.6  \\
OVMM (RL)&  &32.4\%  & 15.6\% &  9.2\% & 1.2\% & 14.6\% & 1132.5  \\
OVMM (Heuristic) &  &30.8\%  &14.4\% &3.6\% &0.8\% &12.4\% &1009.8  \\
 RoboAI & &41.2\% &21.2\% &6.4\% &0.0\% &17.2\% &906.2  \\
 KUZHUM & &55.7\% &50.2\% &35.2\% &11.6\% &38.2\% & 1153.3  \\
 MoManipVLA & &66.1\% &62.6\% &53.1\% &15.8\% &49.4\% &1240.5\\

 \midrule
\end{tabular}
\label{table_1:OVMM}
\vspace{-2mm}
\end{table*}

\begin{table*}[t]
\caption{Ablation experiments for optimization cost terms, where the base method denotes the full setting of our approach. We verify its effectiveness by gradually eliminating the cost term. We further explored the effectiveness of bi-level optimization.}
\centering
\begin{tabular}{@{}lcccccccccc@{}}
\toprule
\multirow{2}{*}{\textbf{Method}}& \multicolumn{2}{c}{\textbf{Search Policy}}
& & \multicolumn{3}{c}{\textbf{Partial Success Rates}} &\multirow{2}{*}{\textbf{\makecell{Overall
\\ SR}}} &\multirow{2}{*}{\textbf{\makecell{Partial
\\ SR}}} &\multirow{2}{*}{\textbf{Step}} &\multirow{2}{*}{\textbf{Latency}} \\
 \cmidrule{2-3}\cmidrule{5-7} 
 & Bi-level & Direct & & FindObj & Pick & FindRex  \\
 \midrule
Base Method & \checkmark  & & &66.1\% &62.6\% &53.1\% &15.8\% &49.4\% &1240.5 &693.1\\
Base Method & & \checkmark & &65.9\% &61.1\% &52.5\% &14.1\% &48.5\% &1373.3 &742.9 \\
Base Method w/o GT. &\checkmark &  &  &23.7\%  &12.7\% & 7.1\% & 1.7\% & 11.3\% & 1187.8 & 737.6 \\
\midrule
w/o Reachability & \checkmark &  &  &65.8\% &61.3\%  & 52.6\% &  13.1\% & 48.2\% & 1174.7 & 682.7 \\
w/o Smoothness & \checkmark       &  &  &66.1\% & 61.7\% &53.1\% &13.8\% &48.7\% & 1185.3 & 688.2 \\
w/o Collision & \checkmark        &  &    &66.3\%  &61.4\% &52.7\% & 15.3\% & 48.9\% & 1174.2 & 692.1 \\
 \midrule
\end{tabular}
\label{table_1:ab}
\vspace{-4mm}
\end{table*}

\subsection{Datasets and Implementation Details}
The Open Vocabulary Mobile Manipulation (OVMM) benchmark \cite{yenamandra2023homerobot} contains 60 scene models approximating the layout of a real house, along with over 18k 3D models of everyday objects.
The mobile manipulation task in the OVMM benchmark is formally defined as “Move a target object from container A to container B”, where the target object is a small object that can be grasped by Hello Robot Stretch \cite{kemp2022design}.
The robot is initialized in an unknown environment and needs to perform "Nav to Recp-A, Gaze, Pick Object, Nav to Recp-B, and Place" stages sequentially to complete the mobile manipulation, where errors in any of these stages will result in a failed manipulation.
We collect mobile manipulation expert trajectories with the heuristic baseline provided by OVMM to fine-tune the off-the-shelf VLA models to bridge the cross-embodied gaps.
We employ OpenVLA-7B to generate fine-grained interaction trajectories. Specifically, each expert trajectory consists of a series of tuples containing visual perception, robotic states, and execution actions. 
We collect 200 pick-and-place demonstration episodes and efficiently fine-tune 10K epochs with LORA on 4 RTX 3090 GPUs.
For the trajectory optimization framework, we employ double annealing to search for physically feasible trajectories between waypoints. 
The number of intermediate steps is calculated based on the position and rotation changes between waypoints with the specified step size (0.05).
The scene ESDF is constructed with 3D object models.
We follow the OVMM benchmark setting and utilize the open vocabulary instance segmentation model as the visual perception module.
The cost hyperparameters $\lambda_1$, $\lambda_2$ and $\lambda_3$ in the optimization objective are set to 10.0, 1.0 and 0.6, respectively.
The safety threshold $\epsilon_0$ in the collision cost is set to 0.1.

\begin{figure*}[t]
  \centering
   \includegraphics[width=1.0\linewidth]{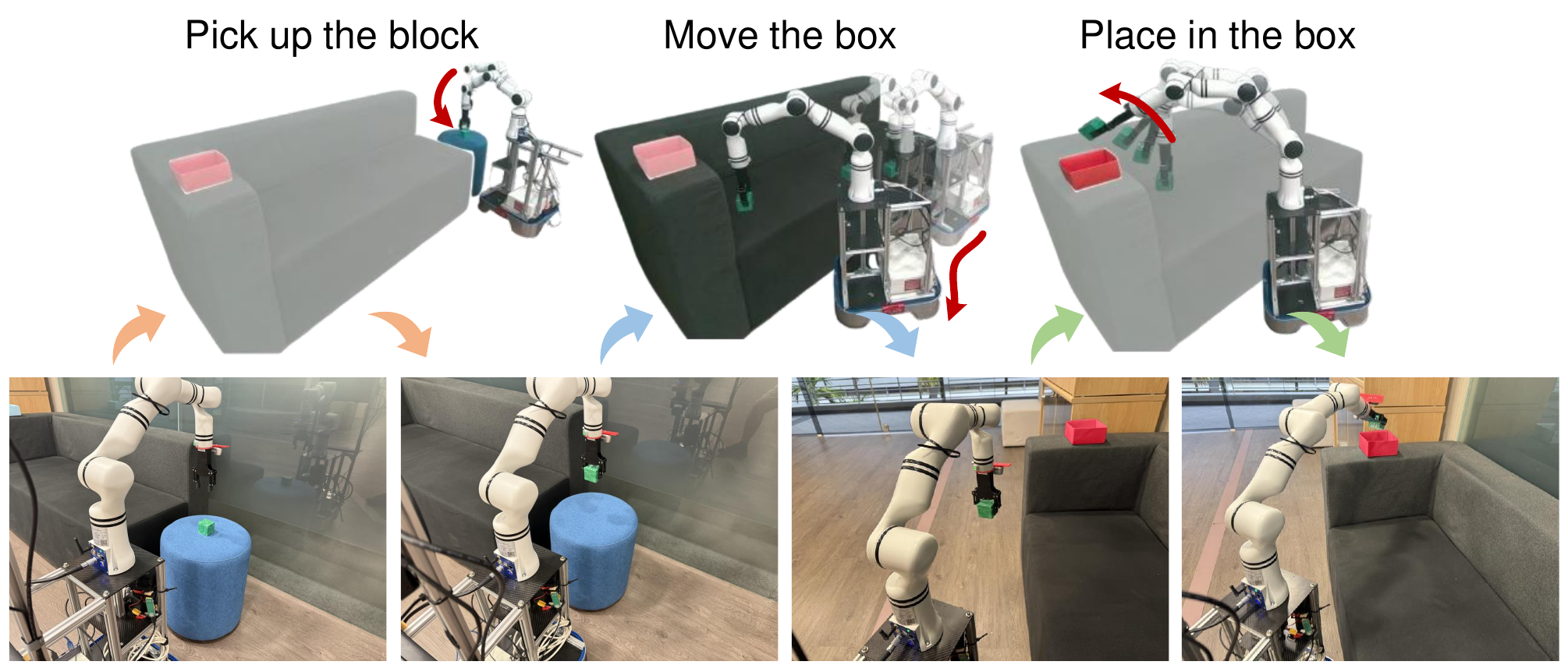}
   \caption{Real-world mobile manipulation visualization results. 
   We demonstrate a real-world mobile robotic Pick-and-Place action sequence, which generates mobile manipulation trajectories through iterative bi-layer search of the base and arm.}
   \label{fig:realworld_vis}
   \vspace{-4mm}
\end{figure*}

\subsection{Comparison with State-of-the-art Methods}
Because our proposed MoManipVLA relies on the VLA models for waypoint guidance, we generate fine-grained base and arm trajectories for the OVMM task exclusively at the end of the agents’ navigation phase within MoManipVLA.
Table \ref{table_1:OVMM} demonstrates the performance of MoManipVLA on OVMM compared to state-of-the-art methods.
Our method achieves 4.2\%  overall success rate and 11.2\% partial
 success rate gain, respectively.
We demonstrate that our method can coordinate the motion of the robotic base and arm in order to keep the end-effector in a reasonable spatial relationship with the target object.
Notably, benefiting from the generalization ability of the pre-trained VLA models, the Pick success rates of our method are 12.4\%  higher than the SOTA methods, which efficiently verify that the proposed approach efficiently transfers the pre-trained VLA model policies to the mobile manipulation task.
Table \ref{table_1:OVMM} demonstrates the efficiency of our approach compared to the baseline, Latency denotes the inference latency for generating the interaction policy.
Notably, the step numbers of our method are close to the heuristic method while the latency is close to the RL-based method, further demonstrating the efficiency of the proposed policy transfer and hierarchical optimization framework.
We further counted the failure cases of mobile manipulation as illustrated in Table \ref{tab:failure_case}. "Find\_recep" means that the grasp receptacle is not found, "Nav\_to\_place" means that the placement receptacle is not found, and "Orient\_to\_place" means that the robot's navigation strategy fails (it is not aligned to the target), further validating the importance of the base waypoints in mobile manipulation.

\subsection{Ablation Experiment}
To fully verify the importance of each cost term in trajectory optimization, we conduct ablation experiments on the baseline method. We respectively remove the terms of reachability, smoothness and collision and evaluate the success rate and the efficiency. 
The results of the ablation experiments are illustrated in Table \ref{table_1:ab}, where “Base Method" denotes our proposal method,  “w/o GT" means using the instance segmentation.
Each term improves the overall success rate with the obvious contribution. Among all three terms, reachability increases the success rate most significantly because mobile manipulation requires the target interaction in large regions, where it is challenging for the agent to reach the target due to the base location. In the mobile manipulation task, the cooperative control of the mobile base and the arm to ensure end-effector reachability is a key bottleneck in performance, as well as the transfer of the pre-trained VLA model to mobile manipulation. 

Meanwhile, we also leverage different pose search methods for trajectory generation. Besides the bi-level objective optimization in our method, we also use the ordinary Dual Annealing search algorithm without search space decomposition. Despite higher success rate, our bi-level objective optimization method also reduces the latency significantly. This indicates the effectiveness of our algorithm to avoid search deficiency.
We further leverage the groundtruth object masks as the visual input for our system to fairly evaluate our mobile manipulation policy. We also evaluate our method by utilizing vision foundation models such as Detic for object mask generation. As the objects are usually highly cluttered in the household environments, the quality of visual perception degrades significantly. Therefore, the overall success rate is far below methods using groundtruth masks. 

\begin{table}[t]
    \centering
    \setlength{\tabcolsep}{5.8pt}
    \caption{Failure cases in OVMM simulation environment}
    \vspace{-2mm}
    \begin{tabular}{c|c|c|c}
     \midrule
        Find\_recep & Nav\_to\_place & Orient\_to\_place & Other \\  \midrule
        14.62\% & 0.33\% & 72.09\% & 12.96\% \\  \midrule
    \end{tabular}
    \label{tab:failure_case}
    \vspace{-5mm}
\end{table}

\subsection{Real World Experiment}
Table \ref{tb:real_world} demonstrates the performance of our approach on real-world robotics. 
We employ the hexman echo plus base and RM65 robotic arm components offline mobile platforms and utilize nvblox to reconstruct the scene ESDF and Grounding SAM \cite{ren2024grounded} to obtain masks for the robotic arm and target object, using the arm mask as the query point and excluding the target object mask in ESDF construction.
We follow the ORB-SLAM setting and use the Intel Realsense T265 tracking camera to acquire the real-time camera pose and base pose.
Benefiting from the generalization of the pre-trained VLA models, our approach requires only 50 samples to complete the fine-tuning and achieves 40\% success rate on the mobile manipulation task.
The drawer opening task is a challenging hinge object interaction in mobile manipulation, requiring the robotic trajectory to meet physical constraints and avoid collision with hinge object movement, which leads to 10\% success rate.

\begin{table}[t]
\centering
\caption{Real-world results. We tested each task 10 times.}
\vspace{-1mm}
\setlength{\tabcolsep}{12pt}
\begin{tabular}{cccc}
\midrule
\textbf{Task} & \textbf{SR} & \textbf{Step} & \textbf{Latency(ms)} \\
\midrule
Stack Block & 30.0\% & 67.0 & 580.0 \\
Open Drawer   & 10.0\%      &  89.0     & 592.0      \\
Put in Bowl &  40.0\%     &   102.0    &   585.0    \\
\midrule
\end{tabular}
\label{tb:real_world}
\vspace{-8mm}
\end{table}

We demonstrate the qualitative results in Figure \ref{fig:realworld_vis}, which illustrates the sequence of the Pick-and-Place task of the mobile manipulation. 
After the robot approaches the target object by navigation, the motion of the mobile base and arm is generated based on the waypoints produced by the pre-trained VLA models, ensuring that the end-effector follows the VLA model guidelines.
Initially, target blocks are positioned within the robotic arm's native workspace, enabling task completion through direct execution of VLA models predicted waypoints via generated arm trajectories. 
When the agent needs to move to place the red box, the motion planning at this point is costly and the base needs to be moved to augment the motion space of the arm to indeed reach the waypoints.
We initialize base optimization with heuristic navigation, then refine the base waypoint using bi-level optimization, so that the red box reenters the workspace to complete placement.
Our proposed MoManipVLA tightly integrates navigation and manipulation, ensuring efficient execution of mobile manipulation tasks, and further leverages the generalization ability of the pre-trained VLA models.

\section{Conclusion}
In this paper, we have presented an efficient policy transfer framework to generalize pre-trained VLA models to mobile manipulation tasks. We first predict the waypoints for the end-effectors with the fixed-base VLA models, and generate the trajectories for the mobile base and the robot arm with maximized physical feasibility. By designing objectives considering the end-effector reachability, trajectory smoothness and collision avoidance, we leverage an efficient bi-level objective optimization framework to jointly search for the optimal trajectory pose for the base and the arm. Extensive experiments demonstrate the powerful generalization of our approach in both simulation and the real-world tasks. The limitations of our method are two-fold. First, our proposed method is limited to the performance of the VLA models and can only be deployed in highly constrained movement space.
Second, our system cannot generate effective motions for long-horizon tasks due to the lack of task planning modules. Designing more efficient trajectory generation framework and integrating task planning from foundation models will be our future works.

\section*{Acknowledgement}
This work was supported in part by the National Natural Science Foundation of China under Grant 62376032, in part by the Nanyang Technological University Start up Grant 024303-00001.

{
    \small
    \bibliographystyle{ieeenat_fullname}
    \bibliography{camera_ready}
}


\end{document}